\pdfoutput=1

\documentclass[11pt]{article}

\usepackage[]{EMNLP2023}

\usepackage{times}
\usepackage{latexsym}

\usepackage[T1]{fontenc}

\usepackage[utf8]{inputenc}

\usepackage{microtype}

\usepackage{inconsolata}

\usepackage{algorithm} 
\usepackage{algorithmic} 
\usepackage{amsmath} 
\usepackage{graphicx} 
\usepackage{amssymb}
\usepackage{pifont}
\usepackage{bbm}
\usepackage{xspace}
\usepackage{booktabs}
\usepackage{enumitem}

\newcommand{\sys}{\textit{advICL}\xspace}

\title{Adversarial Demonstration Attacks on Large Language Models}
\author{
Jiongxiao Wang\thanks{\;The first two authors contributed equally.} $^{\ \, 1}$ \quad Zichen Liu$^{2, *}$ \quad Keun Hee Park$^{2}$ \quad Zhuojun Jiang$^{3}$ \\
\textbf{Zhaoheng Zheng}$^{4}$ \quad \textbf{Zhuofeng Wu}$^{5}$ \quad \textbf{Muhao Chen}$^{6}$ \quad \textbf{Chaowei Xiao}$^{1}$\\
\textsuperscript{1}University of Wisconsin-Madison; \textsuperscript{2}Arizona State University;\\ 
\textsuperscript{3}Tokyo Metropolitan University;  \textsuperscript{4}University of Southern California;\\
\textsuperscript{5}University of Michigan - Ann Arbor; \textsuperscript{6}University of California, Davis
  }

\begin{document}
\maketitle
\begin{abstract}
With the emergence of more powerful large language models (LLMs), such as ChatGPT and GPT-4, in-context learning (ICL) has gained significant prominence in leveraging these models for specific tasks by utilizing data-label pairs as precondition prompts. While incorporating demonstrations can greatly enhance the performance of LLMs across various tasks, it may introduce a new security concern: attackers can manipulate only the demonstrations without changing the input to perform an attack. In this paper, we investigate the security concern of ICL from an adversarial perspective, focusing on the impact of demonstrations. We propose a novel attack method named \sys{}, which aims to manipulate only the demonstration without changing the input to mislead the models.  
Our results demonstrate that as the number of demonstrations increases, the robustness of in-context learning would decrease. Additionally, we also identify the intrinsic property of the demonstrations is that they can be used (prepended) with different inputs. As a result, it introduces a more practical threat model in which an attacker can attack the test input example even without knowing and manipulating it. To achieve it, we propose the transferable version of \sys{}, named Transferable-\sys{}. Our experiment shows that the adversarial demonstration generated by Transferable-\sys{} can successfully attack the unseen test input examples. 
We hope that our study reveals the critical security risks associated with ICL and underscores the need for extensive research on the robustness of ICL, particularly given its increasing significance in the advancement of LLMs.

\end{abstract}

\section{Introduction}

The development of large language models (LLMs; \citealt{brown2020language}) has introduced a new paradigm for solving various (NLP) tasks through in-context learning \citep{dong2022survey}. As a novel form of prompt engineering \citep{liu2023pre}, in-context learning prepends providing \emph{demonstrations} (also called \emph{in-context examples}) 
to the test examples as a part of the prompt, so as to assist LLMs in achieving better inference performance for various tasks. 
While numerous studies \citep{xie2021explanation,dai2022can} have demonstrated the efficacy of 
few-shot adaptation of language models with in-context learning, there remains uncertainty regarding potential security risks associated with the usage of demonstrations.

To investigate the security threat of the model, 
numerous adversarial attacks have been developed \citep{ebrahimi2017hotflip, gao2018black, jin2020bert, li2018textbugger, ribeiro2020beyond, li2020bert} with the aim of evaluating the adversarial (worst-case) robustness of the model. 
To easily perform and implement these attacks, TextAttack \citep{morris2020textattack} has been proposed, providing a unified framework.
Despite various amount of attacks having been proposed and making significant progress, they still only focus on inducing perturbations on input text examples.
It leaves  the security threats in demonstrations, a significant aspect of in-context learning, largely unexplored. 
Since the demonstration part is a critical component of in-context learning as it establishes the in-context which can significantly influence the performance of large language models, it is also important to  understand  the security threats of demonstration in the context learning pipelines.

\begin{figure*}[t]
    \centering
    \includegraphics[width=1.0\textwidth]{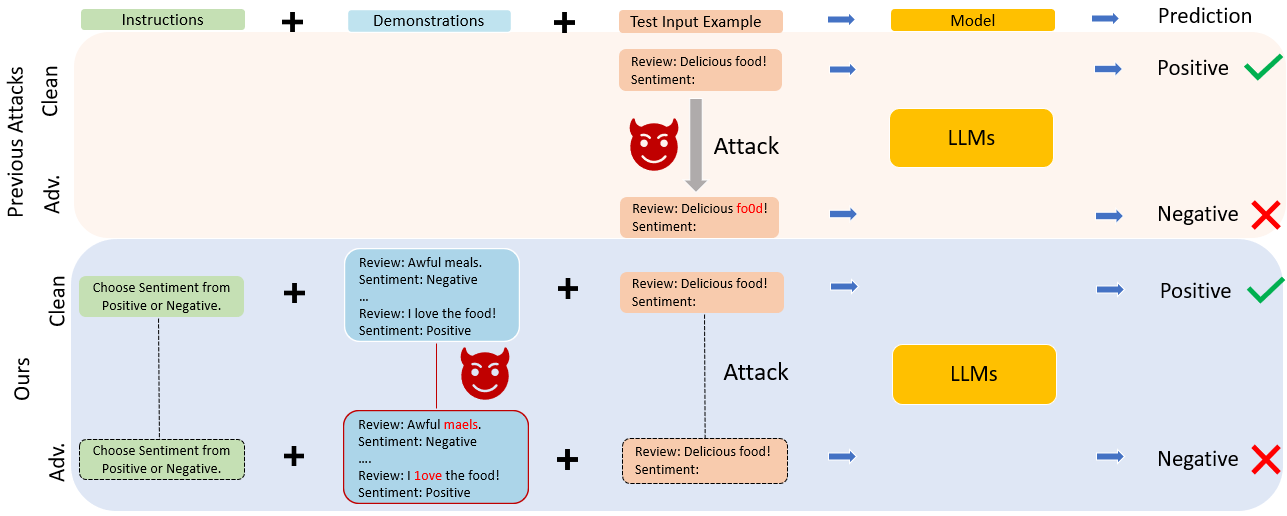}
    \caption{Difference between existing attacks and ours on sentimental classification task. The previous attacks mainly perform attack on the text input examples while ours focuses exclusively on attacking demonstrations.}
    \label{fig:ourvsexisting}
\end{figure*}

In this paper,  we propose a simple yet effective in-context learning 
attack method named \sys to investigate the impact of demonstrations. 
As shown in Figure~\ref{fig:ourvsexisting},  in contrast to standard attacks, which only manipulate the  input text example to perform the attack, \sys{} focuses exclusively on attacking preconditioned demonstrations without manipulating the input text example. 
Specifically, to make our attack easily and flexibly deployable in existing systems, we design our attack under the TextAttack framework but add extra demonstration masking that only allows manipulating demonstration.
Moreover, different from text example-based attacks, given the extended length of in-context learning prompts, standard global similarity constraints \citep{jin2020bert} between the adversarial text and the original text have proven to be less effective, potentially compromising the quality of adversarial examples. To address this challenge, we introduce a demonstration-specific similarity method that applies constraints on each individual demonstration. This method ensures the generation of effective and high-quality adversarial examples in our attack strategy.

We conduct comprehensive experiments on four datasets including SST-2 \citep{socher2013recursive}, TREC \citep{voorhees2000building},  DBpedia \citep{zhang2015character}, RTE \citep{dagan2005pascal} and on diverse LLMs including the GPT2-XL \citep{radford2019language}, LLAMA \citep{touvron2023llama}, Vicuna \citep{vicuna2023}. 
Our method can successfully attack LLMs (e.g. 97.72\% attack success rate (ASR) for LLaMA-7B on DBpedia),  by only manipulating the demonstration without changing the input text. 
We also conduct experiments with our attack method on different numbers of demonstrations (in-context few-shot learning setting). 
Our results demonstrate that although a larger number of demonstrations can potentially increase the performance of LLMs, a larger number of adversarial demonstrations 
are prone to have more threats to the robustness of in-context learning. For instance, implementing \sys{} on the LLaMA-7B model with the DBpedia dataset achieves an ASR of 97.72\% for 8 shots. This is a significant improvement compared to the 1 shot setting, which only reaches 59.39\% ASR under the same condition.


Additionally, one  intrinsic property of the demonstrations is that they can be used (prepended) with different inputs. As a result, it introduces a more practical threat model in which an attacker can attack the test input example even without knowing and manipulating it. To achieve it, we propose the transferable version of \sys{}, named Transferable-\sys{}. 
In detail, Transferable-\sys{} first iteratively and randomly chooses a small set of text examples. It then employs our original \sys{} to generate the ``universally'' adversarial demonstrations that can mislead the model by concatenating with all the selected texts.
Through Transferable-\sys{}, we can reach 72.32\% ASR given different unseen input text examples on DBpedia.

\section{Related work}


\textbf{In-Context Learning.} Current studies on the robustness of in-context learning primarily focus on the instability of demonstrations \citep{liu2023pre}. Researchers have identified various factors that can significantly impact the performance of in-context learning, including demonstration selection \citep{liu2021makes}, demonstration ordering \citep{lu2021fantastically}, and even 
label selection for
demonstrations \citep{wei2023larger}. 
Several methods \citep{wu2022self, liu2021makes, zhao2021calibrate, chen2022relation} have been proposed to improve the stability of in-context learning. For instance, \citet{liu2021makes} present a simple yet effective retrieval-based method that selects the most semantically similar examples as demonstrations, leading to improved accuracy and higher stability. Another work by \citet{zhao2021calibrate} claims that the instability of in-context learning is coming from the language model's bias toward predicting specific answers. Then they propose a method to mitigate this bias by applying an affine transformation \citep{platt1999probabilistic} to the original probabilities, where the transformation weights are computed based on a context-free test, such as ``N/A''. Despite these efforts, the adversarial robustness of in-context learning, particularly when considering perturbations on demonstrations, still remains unexplored, which is exactly the focus of this work.

\noindent\textbf{Adversarial Examples.} The vulnerability of deep neural networks to adversarial examples was first introduced in \citep{goodfellow2014explaining} on image classification tasks. They demonstrated that small perturbations could manipulate the model into incorrect predictions. This concept was later extended to textual data in the field of natural language processing by \citep{papernot2016crafting}, paving the way for various studies in text adversarial attack. The discrete nature of textual examples allows adversarial perturbations to present at three distinct levels: character-level \citep{gao2018black, pruthi2019combating, li2018textbugger}, word-level \citep{jia2019certified, zang-etal-2020-word, ren-etal-2019-generating} and sentence-level \citep{naik2018stress, ribeiro2020beyond}. In this paper, we focus solely on character-level and word-level perturbations.

\section{Method}

We now describe the proposed \sys method starting with the background of in-context learning, followed by technical details of the demonstration attack.

\subsection{Background}
\textbf{In-Context Learning.}
Formally, in-context learning is defined as a conditional text generation problem \citep{liu2021makes}. Given a language model $f$, we aim to generate output $y_{test}$ based on the input test example $x_{test}$ and a demonstration set $C$, where $C$ contains $N$ concatenated data-label pairs $(x_i, y_i)$ with a specific template $s$, noted as $C=\{s(x_1, y_1), ..., s(x_N, y_N)\}$. We can also prepend an optimal task instruction $I$ \citep{dong2022survey} to demonstrations as 
\begin{equation}\label{equ1}
C=\{I, s(x_1, y_1), ..., s(x_N, y_N)\} .
\end{equation}
Then, given the input test example, we can generate $y_{test}$:
\begin{equation}\label{equ22}
y_{test} = f_{generate}(\{C, s(x_{test}, \_)\}) ,
\end{equation}
where $s(x_{test}, \_)$ indicates using the same template as demonstrations but with the label empty.

To apply it to the standard classification tasks, we let the language model choose answer $y_{test}$ from a candidate classification label set $Y=\{c_1, ..., c_k\}$, where $c_k$ is denoted as the label of the $k$-th class. For convenience, we can define a verbalizer function $\mathcal{V}$ \citep{kim2022self} which maps each of the original labels $c_k$ to a specific token $\mathcal{V}(c_k)$. However, not all labels can be directly mapped to a single token. For instance, the label ``Negative'' is composed of two tokens ``Neg'' and ``ative'' when using the GPT2 tokenizer. 
In this case, we define the function $\mathcal{V}$ mapping ``Neg'' and ``ative'' to ``\{space\}Negative'', where an additional space prepended to the word ``Negative''.

To compute the logits $z_k$ of the token $\mathcal{V}(c_k)$ from causal language models as the probability of $c_k$, we can use the following equation:
\begin{equation}
z_k=f_{causal}(\mathcal{V}(c_k)|\{C, s(x_{test}, \_)\}).
\end{equation}
The final prediction result $y_{test}$ is the label with the highest logits probability:
\begin{equation}
\label{equ2}
y_{test} = \mathop{\arg\max}\limits_{c_k \in Y}f_{causal}(\mathcal{V}(c_k)|\{C, s(x_{test}, \_)\}).
\end{equation} 

\smallskip


\subsection{\sys{}} 

For in-context learning, as mentioned before, it consists of both demonstration $C$ and input test examples $x_{test}$, providing a wider attack vector compared to standard adversarial attacks. In this paper, we aim to only manipulate the demonstration $C$ without changing the input text examples $x$ to mislead the models.  
Since the main goal of this paper is to investigate the adversarial robustness of the attack vector on demonstrations, 
for simplification, we mainly focus on applying word-level and character-level perturbation following TextBugger \citep{li2018textbugger}.\footnote{ Since our attack is a general method, it can be easily extended to be applied with other types of perturbation. We leave that as the future work.} 

Our attack, \sys{}, builds upon TextAttack \citep{morris2020textattack}, a standard attack framework. However, unlike other adversarial text attack methods, which manipulate the input test example  $x_{test}$ by adding the perturbation $\Delta$, here we add a mask to only manipulate the demonstration $C$. 
Additionally, since the demonstration $C$ consists of multiple sentence label pairs (($x_1, y_1), ..., (x_N, y_N$)), we sequentially set individual perturbation bounds $\Delta_i$ for each $x_i$ using cosine similarity, which can be computed by the following equation: 
\begin{equation}
CosSim(x_i,x_i') = \frac{\textbf{e}(x_i) \cdot \textbf{e}(x_i')}{||\textbf{e}(x_i)||\;||\textbf{e}(x_i')||} 
\end{equation}
where $x_i'$ is the perturbed sentence and $\textbf{e}(\cdot)$ represents the embedding vector computed by Universal Sentence Encoder \citep{cer2018universal}.

This approach allows us to control the perturbations specifically for each demonstration, providing more fine-grained control over the attack process. Within this context,  we can formalize our objective function: 

\begin{equation}
\mathop{\min}\limits_{\delta \in \Delta}\mathcal{L}(f(\{C_{\delta}, s(x_{test}, \_)\}), y_{test}) ,
\end{equation}

\noindent
where $\mathcal{L}$ calculates the probability of the ground truth label $y_{test}$ using language model $f$, $C_{\delta} = \{I, s(x_1+\delta_1, y_1), ..., s(x_N+\delta_N, y_N)\}$, $\delta_i$ represents the perturbation added to the demonstration $x_i$ under the given bound $\Delta_i$. 


To solve the above objective function, we consider a more practical black-box setting and adopt the greedy search  in TextAttack framework\citep{morris2020textattack}. Specifically,  given a demonstration set $C$ with a test example pair $(x_{test}, y_{test})$, we initially select words from $x_i$ in $C$ to form a $WordList$ based on the word importance level. For each word in the $WordList$, we compute the objective function value $\mathcal{L}$ by adding the word with 4 types of perturbations, including \textit{Character Insertion}, \textit{Character Deletion}, \textit{Character Swap} and \textit{Word Swap} from TextAattack framework. We then use the perturbation that yields the largest reduction in the objective function values calculated by $\mathcal{L}$ while maintaining the similarity constraint $CosSim(x_i, x_i')$ within the bound $\Delta_i$. The process iterates through the $WordList$ until a successful attack is achieved or fails when reaching the maximum iteration. We also present our pseudo code in Appendix~\ref{appendix:algorithm}.

\section{Transferable-\sys{}}
One intrinsic property of the demonstrations is that they can be used (prepended) with  different inputs. As a result, it introduces a more practical threat model in which an attacker can attack the test input example even without {\em knowing} and {\em manipulating} this example. To achieve such a threat model, one solution is to leverage the transferability of the adversarial demonstration: generating a high transferable adversarial demonstration that can be prepended to an arbitrary example to mislead the model.

However, our \sys{} is only performed on a single specific text example, limiting its potential to be transferable to other test input examples.
To address this limitation, inspired by the concept of Universal Adversarial Perturbation \citep{moosavi2017universal}, previously implemented for image classification tasks, we propose an iterative attack pipeline that aims to generate a universally adversarial demonstration by iteratively attacking a randomly selected set of input examples instead of a single input example, 
denoted as the Transferable-\sys{} (T-\sys{} in short).
This approach can significantly extend the applicability and efficacy of \sys{} across various test examples.
The detailed 
process of the pipeline is shown as follows.

\emph{Step 1:} Given a demonstration set $C$, T-\sys{} randomly identifies a small attack candidate set with $k$ test examples, denoted as $S=\{(x_i, y_i)\}_{i=1}^k$. Note that, to better improve the transferability, here we only randomly select the test example in $S$ with demonstration set $\{C, s(x_i, \_)\}$ that can be precisely classified into $y_i$ under the language model $f$.

\emph{Step 2:}  T-\sys{} randomly shuffles test examples in $S$ to obtain $S_{random}$. The random shuffle here can help us generate optimal demonstrations by preventing stuck in a local minimum.

\emph{Step 3:} 
  Before the iterative attack process, we initialize $C'=C$.
  For each $\{C', s(x_i, \_)\}$ at iteration $i$, where $x_i$ is chosen from $S_{random}$, we apply \sys{} to obtain the adversarial demonstration set $C_i$. Then we update $C'$ with $C_i$ only if $C_i$ has a lower Attack Accuracy than $C'$ on $S$. Note that we need to add an additional constraint that bounds the distance between $C_i$ and $C'$ and the distance between $C_i$ and $C$.
  

We then repeat Step 2 and Step 3 until $R$ rounds are complete.
Our pseudo code of T-\sys{} is also presented in Appendix~\ref{appendix:algorithm}.

\section{Experiment}
In this section, we start by introducing our experimental setup, followed by an evaluation of the effectiveness of \sys{}. We also conduct a perceptual evaluation to ensure the high quality of our generated attack demonstrations. Finally, we carry out ablation studies to evaluate the quality of adversarial demonstrations generated by different similarity constraint methods and the influence of varying templates used in in-context learning.

\begin{table*}[t]
\centering
\resizebox{0.9\textwidth}{!}{
\begin{tabular}{cc|ccc|ccc|ccc|ccc}
\toprule
& \multicolumn{1}{c}{} & \multicolumn{3}{c}{\textbf{DBpedia}}  & \multicolumn{3}{c}{\textbf{SST-2}}  & \multicolumn{3}{c}{\textbf{TREC}}& \multicolumn{3}{c}{\textbf{RTE}}\\
 \textbf{Model}  & \textbf{Metric}    &    1-shot  &  4-shot  &  8-shot     &   1-shot  &  4-shot  &  8-shot     &   1-shot  &  4-shot  &  8-shot &   1-shot  &  4-shot  &  8-shot  \\
\midrule
            & Clean Acc        &  37.6  &  49.4  &  60.4  &  59.4  &  65.6  &  62.6    &  20.0  &  25.2  &  31.4 & 56.8 & 56.8 & 57.8    \\
GPT2-XL     & Attack Acc        &  7.2  &  0.6  &  0.4  &  51.0  &  25.2  &  11.2    &  14.8  &  9.6  &  7.4 & 20.8 & 4.2 & 1.4 \\
            & ASR            &  80.88  &  98.79  &  \textbf{99.38}  &  14.13  &  61.60  &  \textbf{82.15}    &  25.16  &  61.15  &  \textbf{76.12} &63.30 & 92.60 & \textbf{97.70} \\
\midrule
            & Clean Acc        &  71.0  &  78.4  &  77.8  &  69.0  &  81.8    &  92.2  &  53.4  &  51.0  &  58.4  & 73.4 & 59.0 &61.8\\
LLaMA-7B    & Attack Acc        &  28.8  &  6.8  &  1.8  &  40.0  &  45.8  &  40.2    &  31.0  &  11.0  &  5.4 & 16.4 & 0.2 & 0.0 \\
            & ASR            &  59.39  &  91.33  &  \textbf{97.72}  &  41.90  &  44.01  &  \textbf{56.42}    &  41.95  &  78.43  &  \textbf{90.75} & 77.63 & 99.66 & \textbf{100.00}\\
\midrule
            & Clean Acc        & 77.2   &  52.6  & 41.4   & 79.8   &    79.2  & 82.8  & 60.0   &   47.6 & 38.6 &73.0 &69.8&68.2  \\
Vicuna-7B    & Attack Acc        & 40.0   &  6.2  &  1.6  &  57.4  &   43.4 &  27.4   & 40.4  & 10.0  & 8.0 & 30.6 & 5.6 & 0.6 \\
            & ASR            & 48.16  &  88.21  &  \textbf{96.14}  & 28.07   &   45.25 &  \textbf{67.08}  &  32.68  &  79.08  & \textbf{79.40} & 57.90 & 91.93& \textbf{99.11}  \\
\bottomrule
\end{tabular}
}
\caption{Effective of \sys{} among different datasets and models under various shot numbers. The highest ASR value among different shot numbers is highlighted.
}
\vspace{-3mm}
\label{tbl:main}
\end{table*}

\begin{table*}[t]
\centering
\resizebox{0.9\textwidth}{!}{
\begin{tabular}{c|ccc|ccc|ccc|ccc}
\toprule
 \multicolumn{1}{c}{}  & \multicolumn{3}{c}{\textbf{DBpedia}}  & \multicolumn{3}{c}{\textbf{SST-2}}  & \multicolumn{3}{c}{\textbf{TREC}}& \multicolumn{3}{c}{\textbf{RTE}}\\
\textbf{Quality Metric}    &    1-shot  &  4-shot  &  8-shot     &   1-shot  &  4-shot  &  8-shot     &   1-shot  &  4-shot  &  8-shot &   1-shot  &  4-shot  &  8-shot  \\
\midrule

AEQS        & 92.00  & 94.00   & 91.00  & 88.00  & 89.00   & 90.00     & 90.00 & 92.00 & 96.00 & 92.00 & 92.00  & 92.00    \\
CosSim        & 94.65 & 96.52  & 96.78   & 93.80  & 94.76  & 95.25  & 95.94  &  96.80 & 97.79 & 94.09&97.87 & 98.63\\
BLEU           &  95.71  & 96.52  & 97.94   & 92.51  & 91.81  & 92.59   & 97.20  & 96.74  & 97.57 & 87.23 & 93.67 & 95.77\\
\midrule
Adv PPL        &  13.57 & 9.50  & 7.20  & 36.73  & 22.18 & 16.53   & 19.39  & 12.05  & 8.76 & 12.27 & 7.82 & 6.80\\
Original PPL &  12.77 & 6.87  & 5.61  & 46.80  & 14.23 & 10.21   & 21.29  & 9.49  & 6.99 & 8.56 & 6.66 & 6.13\\

\bottomrule
\end{tabular}
}
\caption{
Perceptual Evaluation on 100 successfully attacked examples on LLaMA-7B.
}
\vspace{-3mm}
\label{tbl:quality}
\end{table*}


\subsection{Experimental Setup}
\textbf{Language Models and Dataset.} We conduct experiments using the following language models: the widely used GPT2-XL, recently open-sourced LLaMA-7B, and Vicuna-7B. Vicuna-7B is an instruction-tuned LLaMA-7B with GPT4- or ChatGPT-generated text collected from \url{ShareGPT.com}.

We conduct experiments using four classification datasets, including binary classification datasets SST-2 and RTE, multi-class classification datasets TREC and DBpedia. These datasets cover tasks such as sentiment analysis, textual entailment, topic and question classification tasks. 

\noindent\textbf{In-Context Learning and Attack Settings.}
For in-context learning,  we follow the setting by~\cite{zhao2021calibrate} and use their templates to incorporate demonstrations for prediction. The details of 
these templates are listed in
Appendix~\ref{appendix:template}. 
Regarding the number of demonstrations, we use 1-shot, 4-shot, and 8-shot settings. Specifically, given each test input from 
the test set, we randomly select the demonstration from the training set and repeat the processing 5 times to calculate the average accuracy. 
For attacks, we follow \citet{li2018textbugger} and adopt the untargeted and black-box settings, where we only have access to logits of predicted next tokens. 
To specify the perturbation bounds $\Delta_i$ for our attack method, we employ the cosine similarity 
between the adversarial and original individual demonstration sentences with a threshold of 0.8.

\noindent\textbf{Evaluation Metrics.}
We use Attack Success Rate (ASR) \citep{wang2021adversarial} to evaluate the effectiveness of \sys{}. Given a dataset $\mathcal{D}$ with $N$ data instance $x$ and label $y$,  an adversarial attack method $\mathcal{A}$ that generates adversarial examples $\mathcal{A}(x)$. The ASR is then calculated as
$\textbf{ASR} = \sum\limits_{(x, y)\in \mathcal{D}}\frac{\mathbbm{1}[f(\mathcal{A}(x))\neq y]}{\mathbbm{1}[f(x)=y]}$.
 In addition, we also report the clean accuracy (Clean Acc) which evaluates the classification accuracy with clean demonstrations, and attack accuracy (Attack Acc) which evaluates the classification accuracy with generated adversarial demonstrations.  


\subsection{Effectiveness of \sys{}}
\begin{figure*}
    \centering
    \includegraphics[width=1.0\textwidth]{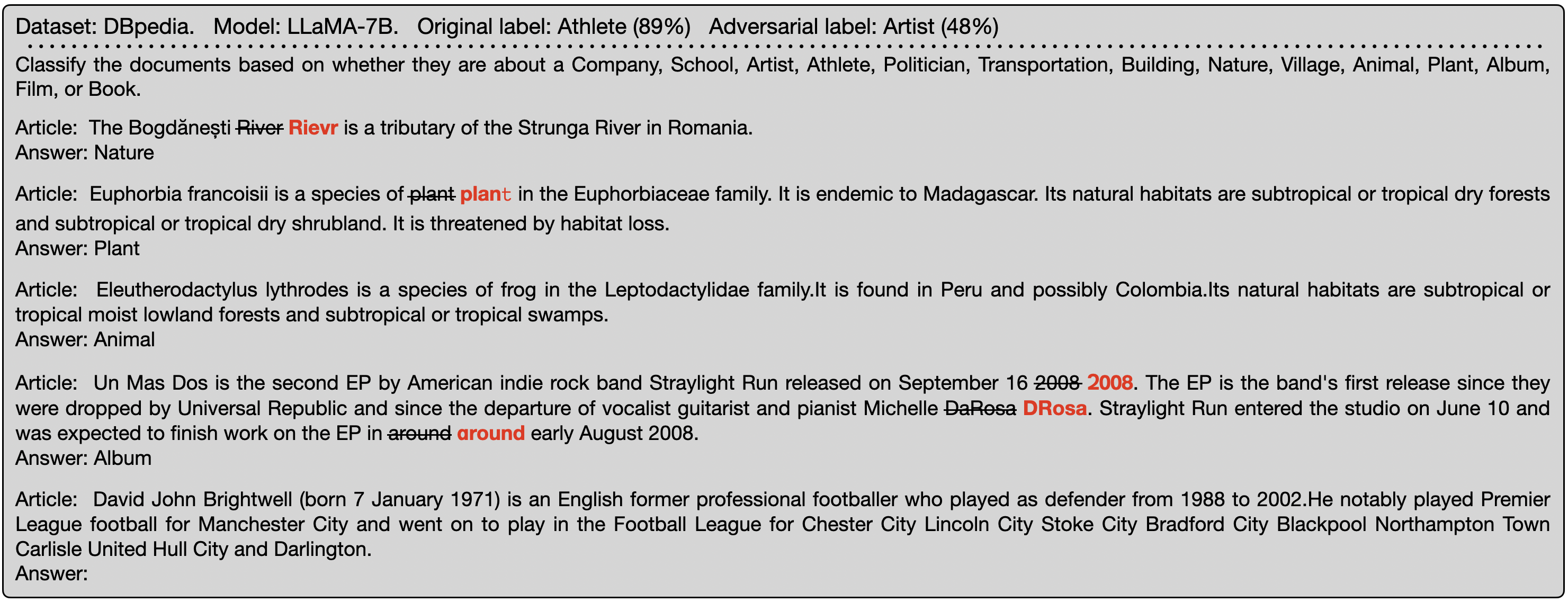}
    \caption{
    Visualization of an adversarial example generated by \sys{} on the DBpeda dataset via attacking  LLaMA-7B model. 
    }
    \label{fig:real_example}
\end{figure*}

Table~\ref{tbl:main} presents the results of \sys, 
from which we can  observe that in-context learning indeed introduces a new security concern. By  manipulating only the demonstrations 
but not the input test examples, attackers can successfully mislead the model. For instance, ASR of \sys{} on DBpedia and RTE with 8-shot demonstrations can reach even higher than 95\% among all test models. We also show the visualization of an adversarial example generated by \sys{} in Figure~\ref{fig:real_example}.

Additionally, 
the consistent trend in Table~\ref{tbl:main} shows that larger shot numbers can bring higher ASR. 
This observation reveals that, although larger shot numbers can normally benefit the performance of in-context learning, they can also introduce a higher potential of threats to LLMs. 

\subsection{Perceptual Evaluation}
To conduct a comprehensive perceptual evaluation of generated adversarial demonstrations, we employ a diverse set of metrics 
that encompass both human evaluation and automatic text quality evaluation.

Our evaluation process begins with a thorough examination of the attacked demonstrations by human annotators. Given a complete set of successfully 
attacked instances with $n$ examples, each instance undergoes a meticulous inspection by the annotators. Evaluations are conducted based on established criteria through three dimensions: semantic coherence, grammatical accuracy, and expression fluency. The annotators then count the number of examples, represented as $c$, that meet the criteria across all three demonstrations. Following this, we calculate the Annotator Evaluation Quality Score (AEQS) using the formula $\text{AEQS}=\frac{c}{n}$.

For automatic text quality evaluation, we employ a selection of metrics: Cosine Similarity (CosSim) and Bilingual Evaluation Understudy (BLEU; \citealt{papineni2002bleu}). These metrics are computed directly from the original and adversarial sentences. In addition, we use Perplexity (PPL) to evaluate the text fluency. Unlike the previous metrics, PPL is calculated for individual sentences. Therefore, we compute perplexity for both the original and adversarial sentences, referred to as Original PPL and Adv PPL respectively. For further details on these machine evaluation metrics, please refer to Appendix~\ref{appendix:metric}.


We present our perceptual evaluation results of \sys{} in Table~\ref{tbl:quality}. For this evaluation, we select 100 successfully attacked examples across all datasets on LLaMA-7B. CosSim, BLEU, Attacked PPL and Original PPL represent the average scores among these 100 examples. Table~\ref{tbl:quality} reveals an average AEQS of 90\% and high machine evaluation scores. Based on these results, we can conclude that our generated adversarial demonstrations maintain high perceptual quality under both human and machine evaluations.


\begin{table}[t]
\centering
\resizebox{0.5\textwidth}{!}{
\begin{tabular}{cc|ccc}
\toprule
 &\multicolumn{1}{c}{}  & \multicolumn{3}{c}{\textbf{SST-2}} \\
\textbf{Attack Method} &\textbf{Quality Metric}    &    1-shot  &  4-shot  &  8-shot   \\
\midrule
&AEQS        & 70.00  & 65.00 &  74.00       \\
 baseline&CosSim        & 91.29  & 92.54   & 92.77  \\
 &Adv PPL        & 41.50 &  24.65   & 18.35  \\
&BLEU           & 92.57  & 90.70  &  91.82 \\
\midrule
&AEQS        & 88.00  & 89.00   &  90.00     \\
 \sys{}&CosSim        & 93.80  & 94.76  &  95.25 \\
 &Adv PPL        &  36.73  & 22.18 & 16.53  \\
&BLEU           &  92.51  & 91.81  &  92.59 \\
\bottomrule
\end{tabular}}
\caption{Similarity constraint analysis among different perceptual quality metrics using the adversarial demonstration generated by attacking LLaMA-7B on the SST-2 dataset. 
}
\vspace{-3mm}
\label{tbl:quality2}
\end{table}

\subsection{Ablation Study}
\noindent\textbf{Similarity constraint.} 
To validate the importance of our new similarity constraint, denoted as $\Delta_i$,  we compare it with the standard global demonstration perturbation bound $\Delta$ used in TextBugger, denoted as the \textit{baseline} in Table~\ref{tbl:quality2} by  perceptual evaluation. We  keep everything the same and just replace the similarity constraint with the standard global demonstration perturbation bound. We use the adversarial demonstrations generated on LLaMA-7B on the SST-2 dataset for analysis. 
The constraint configurations in TextAttck framework for baseline and \sys{} are shown in Appendix~\ref{config}.


As the results presented in Table~\ref{tbl:quality2}, all quality metrics for our method consistently outperform those of baseline, except for the 1-shot BLEU score. This comparison demonstrates that our individual perturbation bounds ensure the high quality of generated adversarial examples. 
We also show visualized examples generated by our \sys{} and baseline respectively in Appendix~\ref{appendix:example}. We can also find that the visual quality of our method is higher than the baseline. 

\noindent\textbf{Different Template.}
A previous study \citep{min2022rethinking} demonstrated that the performance of in-context learning can vary significantly under different templates. To ensure the effectiveness of our attack method across diverse templates, we conducted additional experiments on the SST-2 dataset among different models using an alternative template, as suggested by \citet{min2022rethinking}. This alternative template, referred to as SST-2-Alter, is shown in detail in Appendix~\ref{appendix:template}. The results of these experiments, presented in Table~\ref{tbl:alter}, show high ASR values and consistent increasing trends with larger shots numbers. These results lead us to conclude that our \sys{} performs stably under different templates.



\begin{table}[t]
\centering
\resizebox{0.5\textwidth}{!}{
\begin{tabular}{ccccc}
\toprule
  & & \multicolumn{3}{c}{\textbf{SST-2-Alter}}   \\
 \textbf{Model}      &    \textbf{Metric}         &    1-shot  &  4-shot  &  8-shot        \\
\midrule
            & Clean Acc        &  51.2  &  49.0  &  49.8       \\
GPT2-XL     & Attack Acc        &  26.4  &  9.2  &  2.0    \\
            & ASR            &  48.52  &  81.20  &  \textbf{95.92}    \\
\midrule
            & Clean Acc        &  60.0  &  72.8 &  75.6   \\
LLaMA-7B    & Attack Acc        &  11.4  &  13.4  &  9.8      \\
            & ASR            &  80.98  &  81.56  &  \textbf{87.05}      \\
\midrule
            & Clean Acc        &  58.0  & 54.0   & 55.8    \\
Vicuna-7B    & Attack Acc        & 10.8  & 6.2   &  3.4   \\
            & ASR            & 81.34  & 88.75   &   \textbf{93.87}   \\

\bottomrule
\end{tabular}
}
\caption{
Effectiveness of \sys{} on another different template of SST-2.
}
\label{tbl:alter}
\end{table}



\begin{table*}[t]
\centering
\resizebox{\textwidth}{!}{
\begin{tabular}{cc|ccc|ccc|ccc|ccc}
\toprule
& \multicolumn{1}{c}{} & \multicolumn{3}{c}{\textbf{DBpedia}}  & \multicolumn{3}{c}{\textbf{SST-2}}  & \multicolumn{3}{c}{\textbf{TREC}} & \multicolumn{3}{c}{\textbf{RTE}} \\
 \textbf{Attack Method}  & \textbf{Metric}    &   1-shot  &  4-shot  &  8-shot     & 1-shot  &  4-shot  &  8-shot     & 1-shot  & 4-shot  &  8-shot & 1-shot  & 4-shot  &  8-shot \\
\midrule
            &Avg Clean Acc        &  63.18  &  75.81  &  78.55  &  66.60  &  82.60  &  93.65  &  55.38  &  52.25  &  59.40  & 66.67 & 65.30 &  63.89\\
\midrule
\sys{}    &Avg Attack Acc        &  44.37  &  51.53  &  52.66  &  72.02  &  63.86  &  71.36  &  55.09  &  49.94  &  54.31  &  66.75  &  63.91  &  65.31 \\
    &TASR &   29.77    &  32.03  &  32.96  &   -8.14 &  22.68  &  23.80  &  0.52 & 4.42 &  8.57  &  -0.12  &  2.13  &  -2.22 \\
\midrule
T-\sys{}   &Avg Attack Acc       &  17.09  &  16.87  &  21.74  &  56.30  &  55.53  &  55.30  &  43.38  &  30.99 &  29.65 &  53.30  &  56.27  &  57.72  \\
   &TASR       &  \textbf{72.95}  &  \textbf{77.45}  &  \textbf{72.32}  &  \textbf{15.47}  &  \textbf{32.77}  &  \textbf{40.95}  &  \textbf{21.67}  &  \textbf{40.69} &  \textbf{50.08}  &  \textbf{20.05}  &  \textbf{13.83}  &  \textbf{9.66}  \\
\bottomrule
\end{tabular}
}
\caption{Transferability of adversarial demonstrations generated by T-\sys{} compared with \sys{}. The highest TASR value between two methods is highlighted.}
\vspace{-3mm}
\label{tbl:transfer}
\end{table*}

\subsection{Performance of Transferable-\sys{}}
In this section, we investigate the transferability of adversarial demonstrations generated by \sys{} and Transferable-\sys{}. Specifically, we examine whether an adversarial demonstration generated for an input test example, $x_1$, can be transferred to successfully attack another input test example, $x_2$, even without prior knowledge.

\noindent \textbf{Experimental Setup.}
In order to thoroughly illustrate the transferability of our method, we randomly select $N_{adv}$ adversarial demonstration sets, denoted as $\mathcal{C}$ in total. 
We also randomly sample $N_{test}$ test input examples exclusively from $S$ for transferability evaluation, denoting this test set as $\mathcal{O}$.
For each adversarial demonstration  from $\mathcal{C}$, we evaluate the model performance on all test examples from $\mathcal{O}$. We then average model performance accuracy across the $N_{adv}$ adversarial demonstration sets to obtain the average attack accuracy (Avg Attack Acc). Similarly, the average clean accuracy (Avg Clean Acc) can be calculated by replacing each adversarial demonstration set $C'$ with its corresponding clean demonstration $C$. The formula for computing Avg Attack Acc can be found in the Appendix~\ref{appendix:formulation}. To better evaluate the performance of transferable attack, We also compute the Transfer Attack Success Rate (TASR) for both methods by
\begin{equation}\nonumber
\text{TASR}=\frac{\text{Avg Clean Acc}-\text{Avg Attack Acc}}{\text{Avg Clean Acc}}.
\end{equation}


To evaluate the performance of our T-\sys{}, we conduct attacks for each clean demonstration, we select  a small candidates set $S$ with $k=15$ test example number and $R=3$ iterative rounds.





\noindent\textbf{Transferability Performance.} 
In Table~\ref{tbl:transfer}, we report the Avg Clean Acc, Avg Attack Acc, and TASR for both our \sys{} and T-\sys{}, as applied to the LLaMA-7B model across all four of our datasets. 
From Table~\ref{tbl:transfer}, we found that performing \sys{} only yields a limited TASR, even leading to negative values in some cases. Such negative TASR means that adversarial perturbations generated by \sys{} possess completely no transferability and even enhance the performance of in-context learning.
After applying T-\sys{}, we can see a significant increase of TAST
from 72.32\% to 32.96\% for the DBpedia dataset with 8-shot and negative TASR values would never present in all circumstances.
These results indicate that in contrast to \sys{}, adversarial demonstrations generated by T-\sys{} 
generalize much better and have more stable transferability
 cross various input test examples.

\begin{figure}[h]
    \centering
    \includegraphics[width=0.48\textwidth]{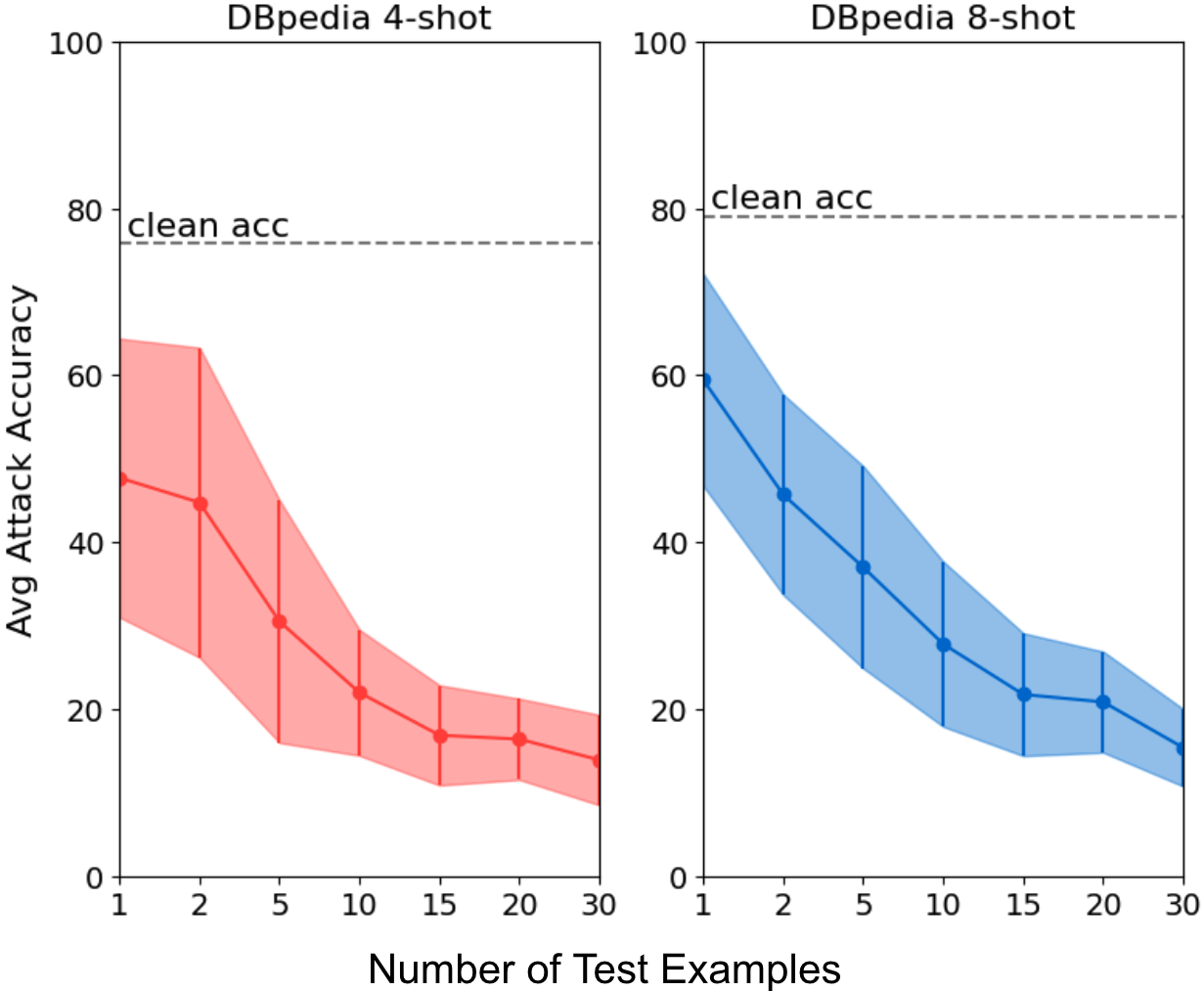}
    \caption{Effectiveness of T-\sys{} with different test example number $k$ of candidates set $S$. The average accuracy is represented by the point, while the shaded area indicates the variance. }
    \label{fig:testset_size}
\end{figure}

\begin{figure}[h]
    \centering
    \includegraphics[width=0.48\textwidth]{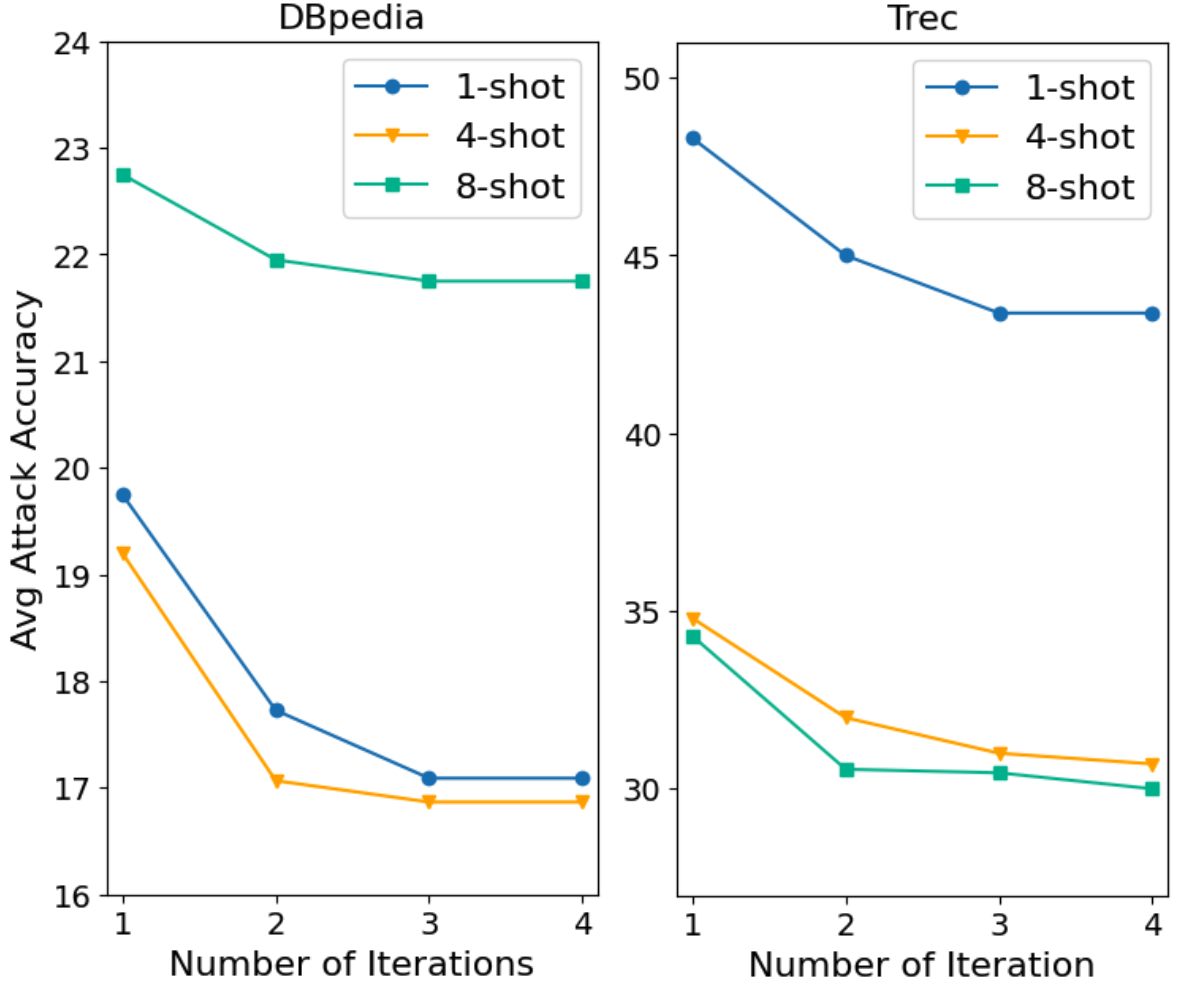}
        \caption{Effectiveness of T-\sys{} among different iteration rounds $R$: Each subplot depicts the average accuracy for three distinct shot numbers.}
    \label{fig:iteration}
\end{figure}

\noindent\textbf{Test Example Number in Candidates Set.} We  show how test example numbers $k$ of candidates set $S$ can affect the transferability of generated demonstrations. We conduct extra experiments on LLaMA-7B with various $k$ and fixed iterative rounds $R=3$ and show their Avg Attack Accuracy and its variance in Figure~\ref{fig:testset_size}.

From Figure~\ref{fig:testset_size}, it is clear that a larger value of $k$ results in a lower Avg Attack Acc, which implies enhanced attack performance. Interestingly, we also observe that as $k$ increases, the variance of Attack Acc diminishes, further indicating that a larger $k$ contributes to the performance stability of transferable demonstrations generated by T-\sys{}.

\noindent\textbf{Iterative Rounds.}
To investigate the role of iterative rounds $R$ of T-\sys{}, additional experiments on LLaMA-7B are conducted with different $R$ and fixed $k=15$. We present our results in Figure~\ref{fig:iteration}.

As seen in Figure~\ref{fig:iteration}, there is a decreasing trend in Avg Attack Acc as the number of iterative rounds increases. It's also observed that the performance of our T-\sys{} tends to converge at around $R=3$ iterative rounds. This suggests that $R=3$ is an optimal number of iterations, with a balance between effectiveness and efficiency.

\section{Conclusion}
This paper studies adversarial robustness of in-context learning, with a particular focus on demonstration attacks. By performing our \sys{}, we find that demonstrations used in in-context learning are vulnerable to adversarial attacks. Even worse, a larger number of demonstrations can exacerbate such security concerns. Additionally, using Transferable-\sys{}, we also demonstrate that our generated adversarial demonstrations tend to be transferrable to a wider range of input text examples such that it can attack the input text examples even without knowing them. 


\section*{Limitations}
To ensure easy and flexible deployment of our attack, we implement it in the TextAttack framework using a black-box approach. However, it has a limitation in terms of time cost. Since we assume no access to the model and cannot calculate gradients through back propagation, we need to estimate the gradient direction to generate adversarial examples, resulting in higher computational requirements compared to other white-box methods.  Since the primary objective of this paper is to demonstrate the new emergent security threats introduced in the in-context learning framework, and the design of our method is not dependent on any specific attack methods, we leave the implementation of a more efficient attack strategy as future work.

\section*{Ethics Statement}

Through the investigation of the in-context learning from the security perspective, we hope our work can raise awareness for the community of such vulnerabilities. We highlight the importance of the demonstration in the context learning framework and inspire the community to design protection strategies for safe storing, retrieving and verifying the demonstration data for making predictions. All data, models we use in this work are publicly available. 
\bibliography{custom}
\bibliographystyle{acl_natbib}

\clearpage

\appendix

\section{Pseudo Code}\label{appendix:algorithm}
\subsection{\sys{}}
Peseudo code of \sys{} is presented in Algorithm~\ref{alg1}. Two extra functions of $WorkImportanceRank$ and $SelectBug$, whose pseudo codes are shown in Algorithm~\ref{alg3} and Algorithm~\ref{alg4} respectively, would be used in Algorithm~\ref{alg1}.

In line 5 of Algorithm~\ref{alg3}, $f_{y_{test}}$ presents the logits value of ground truth label $y_{test}$ using language model $f$ and $C_{adv}/\{w_j\}$ means delete word $w_j$ in demonstration set $C_{adv}$.

In line 2 of Algorithm~\ref{alg4}, function $BugGenerator$ can generate a set named $bugs$ which has four kinds of bugs including \textit{Character Insertion}, \textit{Character Deletion}, \textit{Character Swap} and \textit{Word Swap}.

\subsection{Transferable-\sys{}}
Pseudo code for Transferable-\sys{} is illustrated in Algorithm~\ref{alg2}. It also uses functions of $WorkImportanceRank$ and $SelectBug$.

\begin{algorithm*}[h]
\caption{\sys}
\label{alg1}
\begin{algorithmic}[1]
\REQUIRE{Test example pair $(x_{test}, y_{test})$, demonstration set $C=\{I, s(x_1, y_1), ..., s(x_N, y_N)\}$, language model $f$, similarity threshold $\varepsilon$}
\ENSURE{Adversarial demonstration set $C_{adv}=\{I, s(x_{1_{adv}}, y_1), ..., s(x_{N_{adv}}, y_N)\}$}

\STATE Initialize: $C_{adv} \gets C$, $\mathcal{V} \gets Vocabulary(\{x_1, ..., x_N\})$;
\STATE $WordList_{ordered} = WordImportanceRank(\mathcal{V}, C_{adv}, s(x_{test}, y_{test}), f)$;
\FOR{$w_j$ in $WordList_{ordered}$}
    \STATE $w_j' = SelectBug(w_j, C_{adv}, s(x_{test}, y_{test}), f)$;

    \STATE $x_i' \gets$ replace $w_j \in x_i$ with $w_j'$, $C_{adv} \gets$ replace $x_{i_{adv}} \in C_{adv}$ with $x_i'$;
    \IF{$CosSim(x_i, x_i') \leq \varepsilon$}
        \STATE \textbf{return} None
    \ELSIF{$f(\{C_{adv}, s(x_{test}, \_)\}) \neq y_{test}$}
        \STATE \textbf{return} $C_{adv}$
    \ENDIF
\ENDFOR
\STATE \textbf{return} None
\end{algorithmic}
\end{algorithm*}

\begin{algorithm*}[h]
\caption{Transferable-\sys{}}
\label{alg2}
\begin{algorithmic}[1]
\REQUIRE{Demonstration set $C=\{I, s(x_1, y_1), ..., s(x_N, y_N)\}$, original demonstration set $C_{original}=\{I, s(x_{1_o}, y_1), ..., s(x_{N_o}, y_N)\}$, test examples set $S=\{(x_i, y_i)\}_{i=1}^k$, language model $f$, maximum iteration rounds $MaxIter$, step-wise similarity threshold $\varepsilon$, accumulative similarity threshold $\varepsilon_a$}
\ENSURE{Adversarial demonstration set $C_{adv}=\{I, s(x_{1_{adv}}, y_1), ..., s(x_{N_{adv}}, y_N)\}$}

\STATE Initialize: $C_{adv} \gets C$;
\STATE $iter \gets 0$, $bestASR \gets 1$;
\WHILE{$iter < MaxIter$ and $bestASR > 0$}
    \STATE $S_{random} \gets RandomShuffle(S)$;
    \FOR{$(x_i, y_i)$ in $S_{random}$}
        \STATE Initialize: $C_i \gets C_{adv}$, $\mathcal{V} \gets Vocabulary(\{x_{1_{adv}}, ..., x_{N_{adv}}\})$;
        \STATE $WordList_{ordered} = WordImportanceRank(\mathcal{V}, C_i, s(x_{test}, y_{test}), f)$;
        \FOR{$w_j$ in $WordList_{ordered}$}
            \STATE $w_j' = SelectBug(w_j, C_i, s(x_i, y_i), f)$;
        
            \STATE $x_l' \gets$ replace $w_j \in x_l$ with $w_j'$, $C_i \gets$ replace $x_l \in C_i$ with $x_l'$;
            \IF{$CosSim(x_l, x_l') \leq \varepsilon$ or $CosSim(x_{l_o}, x_l') \leq \varepsilon_a$}
                \STATE Roll back the attack on $w_j$, $C_i \gets$ replace $x_l' \in C_i$ with $x_l$;
                \STATE Break
            \ELSIF{$f(\{C_i, s(x_{test}, \_)\}) \neq y_{test}$}
                \STATE Break
            \ENDIF
        \ENDFOR
        
        \STATE Evaluate ASR of $C_i$ on $S$: $ASR \gets Evaluate(C_i, S)$;
        \IF{$ASR \leq bestASR$}
            \STATE $bestASR \gets ASR$, $C_{adv} \gets C_i$;
        \ENDIF
    \ENDFOR

\ENDWHILE

\STATE \textbf{return} $C_{adv}$

\end{algorithmic}
\end{algorithm*}

\begin{algorithm*}
\caption{Word Importance Ranking}
\label{alg3}
\begin{algorithmic}[1]
\STATE \textbf{function} WordImportanceRank($\mathcal{V}, C, s(x_{test}, y_{test}), f$)
\STATE Initialize: $WordList \gets EmptySet()$;
\FOR{$w_j$ in vocabulary set $\mathcal{V}$}
    \IF{$w_j \in x_i, i \in 1...N$}
        \STATE Compute word importance score $D_{w_j}$ for $w_j$ \\
        $D_{w_j} = f_{y_{test}}(\{C_{adv}, s(x_{test}, \_)\})-f_{y_{test}}(\{C_{adv}/\{w_j\}, s(x_{test}, \_)\})$;
        \STATE $WordList$ append $w_j$;
    \ENDIF
\ENDFOR
\STATE $WordList_{ordered} \gets Sort(WordList)$ with descending $D_{w_j}$;
\STATE \textbf{return} $WordList_{ordered}$
\STATE \textbf{end function}
\end{algorithmic}
\end{algorithm*}

\begin{algorithm*}
\caption{Bug Selection}
\label{alg4}
\begin{algorithmic}[1]
\STATE \textbf{function} SelectBug($w, C, s(x_{test}, y_{test}), f$)
\STATE $bugs \gets BugGenerator(w)$;
\FOR{each $b_k$ in $bugs$}
\STATE $C'_k \gets$ replace $w$ with $b_k$ in $C$;
\STATE $score(k) = f_{y_{test}}(\{C, s(x_{test}, \_)\}) - f_{y_{test}}(\{C'_k, s(x_{test}, \_)\})$;
\ENDFOR
\STATE $bug_{best} \gets \arg\max_{b_k} score(k)$;
\STATE \textbf{return} $bug_{best}$
\STATE \textbf{end function}
\end{algorithmic}
\end{algorithm*}

\section{In-Context Template}\label{appendix:template}
Table~\ref{template} shows templates employed in our study for the DBpedia, SST-2, TREC and RTE datasets. For an alternative template for SST-2, we utilized the minimal template from \citep{min2022rethinking}, denoted as SST-2-Alter.

\begin{table*}[h]
\centering
\resizebox{\linewidth}{!}{%
\begin{tabular}{l|l|l}
\hline
Dataset & Template  & Example\\
\hline
DBpedia& [Instruction] &Classify the documents based on whether they are about a Company, School, Artist, Athlete, Politician, \\
& & Transportation, Building, Nature, Village, Animal, Plant, Album, Film, or Book. \\
&  & \\
& Article: [sentence]& Article: It's a Long Long Way to Tipperary is a 1914 Australian silent film based on the song It's a Long \\
&&Way to Tipperary by Jack Judge.\\
& Answer: [label] & Answer: Film\\
\hline
SST-2& [Instruction] & Choose sentiment from Positive or Negative .\\
&  & \\
& Review: [sentence] & Review: i had to look away - this was god awful .\\
& Sentiment: [label]& Sentiment: Negative\\
\hline
TREC& [Instruction] &Classify the questions based on whether their answer type is a Number, Location, Person, Description, \\
& & Entity, or Abbreviation.\\
&  & \\
& Question: [sentence]& Question: How many trees go into paper making in a year? \\
& Answer Type: [label] & Answer Type: Number\\
\hline
RTE& [Instruction] &\\
&  & \\
&[sentence 1]&At eurodisney, once upon a time is now in this magical kingdom where childhood fantasies and make-
\\
&&believe come to life. \\
&question: [sentence2]&question: EuroDisney is located in this magical kingdom. True or False? \\
& answer: [label] & answer: False\\
\hline
SST-2-Alter& [Instruction] & \\
&  & \\
& [sentence]& i had to look away - this was god awful .\\
& [label] & Negative\\
\hline
\end{tabular}}
\caption{This table presents template designs for all datasets used in this paper. An extra example for each template is provided for better understanding.}
\label{template}
\end{table*}

\section{Evaluation Metrics}\label{appendix:metric}
Here, we provide a brief introduction to three distinct automatic evaluation metrics commonly employed in NLP:

\textbf{Cosine Similarity (CosSim):} To evaluate semantic coherence between original and adversarial sentences, we involve calculating the cosine similarity on the entire adversarial test prompt with the original one.

\textbf{Bilingual Evaluation Understudy (BLEU):} BLEU is a commonly employed method for evaluating structural similarity between two sentences in an automated manner. Here we utilize the BLEU score to compare the original sentence with the adversarial sentence, assessing the degree of similarity in their grammatical structures.


\textbf{Perplexity (PPL):} PPL is a widely used metric in neural language processing which measures the text fluency of language models. In our scenario, we compute perplexity of both the original and adversarial sentences, denoted as Original PPL and Adv PPL, respectively.

\section{TextAttack Configuration}\label{config}
As our \sys{} is built upon the TextAttack framework, we use certain configurations and functions from TextAttack. In Table~\ref{tbl:config}, we list all configurations of \sys{} in comparison to the baseline TextBugger in TextAttack framework. The differences are highlighted using colored text for easy reference.

\begin{table*}[h]
\centering
\renewcommand{\arraystretch}{1.0} 
\setlength{\tabcolsep}{8pt} 
\small
\label{tab:textattack}
\begin{tabular}{p{0.2\linewidth} | p{0.4\linewidth}| p{0.4\linewidth}}
\toprule
\textbf{Attack Algorithm} & \textbf{\sys} & \textbf{TextBugger} \\
\toprule
\textbf{Search Method} & GreedyWordSwapWIR & GreedyWordSwapWIR \\
\midrule
\textbf{WIR Method} & delete & delete \\
\midrule
\textbf{Goal Function} & UntargetedClassification  & UntargetedClassification\\
\midrule
\textbf{Transformation} & CompositeTransformation & CompositeTransformation \\
\midrule
& \textcolor{red}{IclUniversalSentenceEncoder} & \textcolor{red}{UniversalSentenceEncoder} \\
& \quad \small - Metric: angular & \quad \small - Metric: angular \\
& \quad \small - Similarity Const: \textcolor{red}{single-demonstration} & \quad \small - Similarity Const: \textcolor{red}{multi-demonstration} \\
& \quad \small - Threshold: 0.8 & \quad \small - Threshold: 0.8\\
\textbf{Constraints} & \quad \small - Window Size: $\infty$ & \quad \small - Window Size: $\infty$\\
& \quad \small - Skip Text Shorter Than Window: False & \quad \small - Skip Text Shorter Than Window: False\\
& \quad \small - Compare Against Original: True & \quad \small - Compare Against Original: True\\
& RepeatModification & RepeatModification\\
& StopwordModification & StopwordModification\\
& InstructionModification & InstructionModification\\
\midrule
\textbf{Is Black Box} & True & True \\
\bottomrule
\end{tabular}
\caption{Framework Configuration of \sys{} and TextBugger. Highlighted functions show the difference.}
\label{tbl:config}
\end{table*}

\section{Examples Under Different Similarity Constraint}\label{appendix:example}
\begin{figure*}[h]
    \centering
    \includegraphics[width=1.0\textwidth]{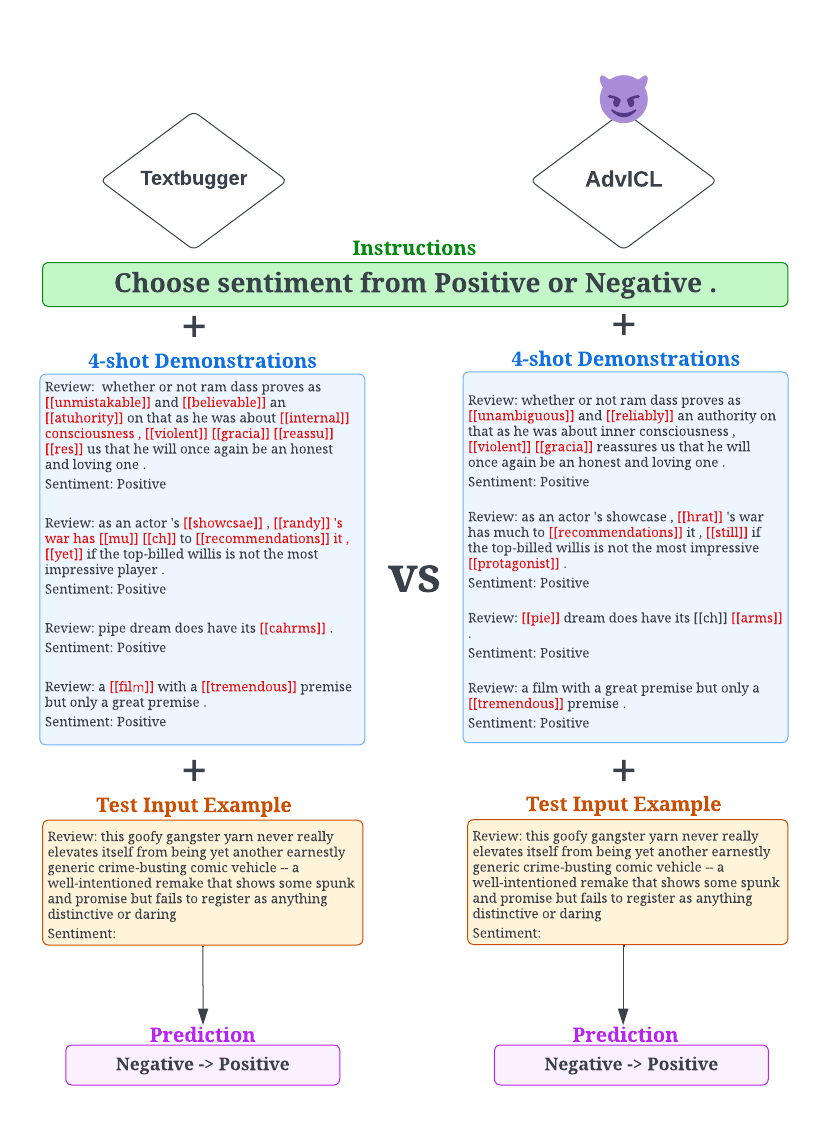}
    \caption{Adversarial demonstrations generated by TextBugger (left) and our \sys{} (right). Perturbed words are all highlighted in red compared with original sentences.}
    \label{fig:enter-label}
\end{figure*}

To offer a clearer perception of the differences in quality between adversarial demonstrations generated by the baseline TextBugger and \sys{}, we present two examples for comparison in Figure~\ref{fig:enter-label}. It is evident from the figure that \sys{} not only performs a successful attack but also generates adversarial demonstrations with a prominent enhancement in quality.

\section{Formulation of Avg Attack Acc}\label{appendix:formulation}
We show the formulation of computing Avg Attack Acc in Figure~\ref{formulation}.

\begin{figure*}[h]
$\text{Avg Attack Acc} = \frac{1}{N_{adv}N_{test}}\sum_{C' \in \mathcal{C}}\sum_{(x_i, y_i) \in \mathcal{O}}\mathbbm{1}[y_i = \mathop{\arg\max}\limits_{c_k \in Y}f_{causal}(\mathcal{V}(c_k)|\{C, s(x_i, \_)\})]$
\caption{Formulation for computing Avg Attack Acc.}
\label{formulation}
\end{figure*}

\end{document}